\documentclass{article}

\usepackage{ijcai2017/ijcai17}
\usepackage{graphicx}
\usepackage{amsthm}
\usepackage{amsmath}
\usepackage{amssymb}
\usepackage{booktabs}
\usepackage{verbatim}
\usepackage{url}
\usepackage{comment}

\usepackage[colorinlistoftodos,disable]{todonotes}

\newcommand{\cis}{$^*$}
\newcommand{\msps}{$^\dagger$}

\excludecomment{short}
\includecomment{extended}

\newcommand{\citeA}[1]{\citeauthor{#1}~[\citeyear{#1}]}

\title{Explainable AI: Beware of Inmates Running the Asylum\\{\small Or: How I Learnt to Stop Worrying and Love the Social and Behavioural Sciences}}
\author{Tim Miller\cis \and Piers Howe\msps \and Liz Sonenberg\cis\\
\cis School of Computing and Information Systems\\
\msps Melbourne School of Psychological Sciences\\
University of Melbourne,  Australia\\
 \{tmiller,pdhowe,l.sonenberg\}@unimelb.edu.au
}

\begin{document}
\maketitle

\begin{abstract}
In his seminal book \emph{The Inmates are Running the Asylum: Why High-Tech Products Drive Us Crazy And How To Restore The Sanity} [2004, Sams Indianapolis, IN, USA], Alan Cooper argues that a major reason why  software is often poorly designed (from a user perspective) is that programmers are in charge of design decisions, rather than interaction designers. As a result, programmers design software for themselves, rather than for their target audience; a phenomenon he refers to as the `\emph{inmates running the asylum}'. This paper argues that explainable AI risks a similar fate. While the re-emergence of explainable AI is positive, this paper argues most of us as AI researchers are building explanatory agents for ourselves, rather than for the intended users. But explainable AI is more likely to succeed if researchers and practitioners understand, adopt, implement, and improve models from the vast and valuable bodies of research in philosophy, psychology, and cognitive science; and if evaluation of these models is focused more on people than on technology. From a light scan of literature, we demonstrate that there is considerable scope to infuse more results from the social and behavioural sciences into explainable AI, and present some key results from these fields that are relevant to explainable AI.
\end{abstract}


\section{Introduction}
\label{sec:intro}

\begin{quote}
 ``\emph{Causal explanation is first and foremost a form of social interaction. One speaks of giving causal explanations, but not attributions, perceptions, comprehensions, categorizations, or memories. The verb to explain is a three-place predicate: \emph{Someone} explains \emph{something} to \emph{someone}. Causal explanation takes the form of conversation and is thus subject to the rules of conversation.}'' --- \citeA{hilton1990conversational}.
\end{quote}

The term ``explainable AI'' has regained traction again recently, after being considered important in the 80s and 90s in expert systems particularly; see \cite{chandrasekaran1989explaining}, \cite{swartout1993explanation}, and \cite{buchanan1984rule}. 
High visibility of the term, sometimes abbreviated XAI, is seen in grant solicitations~\cite{DARPA2016}
and in the popular press~\cite{Nott17}. 
One area of explainable AI receiving attention is explicit \emph{explanation}, on which we say more below.

While the title of the paper is deliberately tongue-in-cheek, the parallels with \citeA{cooper2004inmates} are real: leaving decisions about what constitutes a good explanation of complex decision-making models to the  experts who understand these models the best is likely to result in failure in many cases. Instead, models should be built on an understanding of explanation, and should be evaluated using data from human behavioural studies.

In Section~\ref{sec:survey}, we describe a simple scan of the 23 articles posted as `Related Work' on the workshop web page. We looked at two attributes: whether the papers were built on research from philosophy, psychology, cognitive science, or human factors; and whether the reported evaluations involved human behavioural studies. The outcome of this scan supports the hypothesis that ideas from social sciences and human factors are not sufficiently visible in the field.

In Section~\ref{sec:where-to}, we present some key bodies of work on explanation and related topics from social and behavioural sciences that will be of interest to those in explainable AI, and briefly discuss what their impact could be.


\section{Explainable AI Survey}
\label{sec:survey}

To gather some data to test the hypothesis that the social sciences and human behavioural studies are not having enough impact in explainable AI, a short literature survey was undertaken. This survey is not intended to be even close to comprehensive -- it is merely illustrative. However, the results that it shows are reflective of many other papers in the area that the authors have read.

\subsection{Selected Papers}

The articles surveyed were taken from the `Related Work' list that was posted on the website for the IJCAI 2017 Explainable AI workshop\footnote{See \url{http://home.earthlink.net/~dwaha/research/meetings/ijcai17-xai/}.} as of 16 May 2017 --- the workshop to which this paper is submitted. In total 23 articles were on the list, although one was not included in the results as described later. 
\begin{extended} 
This list can be found in Appendix~\ref{appendix:paper-list}.
\end{extended}

As noted already, this list is far from comprehensive, however, it is a useful list for two reasons:

\begin{enumerate}

 \item First, it was compiled by the explainable AI community: the organisers of the conference requested that people send related papers to be added to the list. As such, it represents at least a subset of what the community see at the moment as highly relevant papers for researchers in explainable AI.

 \item Second, it is objective from the perspective of the authors of this paper. We did not contribute to the list, so the selection is not biased by our argument.
\end{enumerate}

While the authors of some of the listed papers may not consider their work as explainable AI, almost all of the papers were describing methods for automatically generating explanations of some type.

The paper that was excluded is Tania Lombrozo's survey paper on explanation research in cognitive science \cite{lombrozo2012explanation}. This is not an explainable AI paper -- indeed, it summarises one of the bodies of work of which we argue people should be more aware.

\subsection{Survey Method}

The survey was lightweight: it only looked for evidence that the presented research was somehow influenced by a scientific understanding of explanations, and that the evaluations were performed using data derived from human behaviour studies or similar. We categorised the papers on the three items of interest, with the criteria for the scores as follows:

\begin{enumerate}

 \item \emph{On topic}: Each paper was categorised as either being about explainable AI or not, based on our understanding of the topic. It is possible that some papers were included on the workshop website because they presented good challenges or potentially useful approaches, but were not papers about explanation \emph{per se}, in which case they were `off topic'.

 \item \emph{Data Driven}: Each paper was given a score from 0--2 inclusive. 

  A score of 1 was given if and only if one or more of the \emph{references} of the paper was an article on explanation in social science, meaning that: (a) explanation or causal attribution as done by humans is one of the main topics of the referenced article(s); (b) the referenced article(s) validated their claims using data collected from human behaviour experiments; and (c) the referenced article(s) appear in a non-computer science venue \emph{or} in a computer science venue but contributed to the understanding of explanation in general (outside of AI).

A score of 2 was given if and only if (a), (b), and (c) above held, and the survey article (not the referenced article) described an algorithm for automatically generating explanations and this algorithm was derived from data from the social sciences. In other words, the algorithm is explicitly based on a model from one or more of the references.

 A score of 0 was given for any other paper; that is, no references satisfying (a), (b), and (c).

 \item \emph{Validation}: Each paper was given a binary 0/1. A score of 1 was given if and only if the evaluation in the survey article (note, not the referenced article) was based on data from human behavioural studies. Even if the algorithm is categorised as data driven, we argue that it is still important to test that the assumptions and trade-offs made are suitable. It is therefore necessary to (eventually) perform behavioural studies to test if the explanations produced by the algorithm are appropriate for humans. 

\end{enumerate}

\subsection{Results}

Table~\ref{tab:results} shows the results for the survey. 
\begin{extended}
Results for each of the surveyed articles are available in Appendix~\ref{appendix:data}.
\end{extended}
Five papers were deemed `off topic', however, the results are included because we could not know the intent of  those who submitted articles to the reading list.
For the `Data driven' entry, column `N' means that we were unsure about the reference. In this case, one paper had a reference to a cognitive science article of which we were unable to locate a copy. For the `Validation' entry, column `N' means `not applicable': three papers were categorised as not applicable because their status were not research articles, but review articles or position papers, and thus, they did not present any algorithm or model to evaluate. 

\begin{table}[!ht]
\centering
\begin{tabular}{llrrrrrrrr}
 \toprule
                   & \multicolumn{4}{c}{\textbf{On topic}} & \multicolumn{4}{c}{\textbf{Off topic}}\\
                   & \multicolumn{4}{c}{(17 articles)} & \multicolumn{4}{c}{(5 articles)}\\
\cmidrule(lr){2-5}\cmidrule(lr){6-9}
\textbf{Criterion} & N & 0 & 1 & 2 & N & 0 & 1 & 2 \\
\midrule
Data driven    & 1 & 11 & 4 & 1 & 0 & 4 & 0 & 1\\
Validation  & 3 & 10 & 4 & --- & 0 & 4 & 1 & ---\\
\bottomrule
\end{tabular}
\caption{Results on small survey}
\label{tab:results}
\end{table}

These results show that for the on-topic papers, only four articles referenced relevant social science research, and only one of them truly built a model on this. Further, serious human behavioural experiments are not currently being undertaken. For off topic papers, the results are similar: limited input from social sciences and limited human behavioural experiments.

\subsection{Discussion}

The results, while only on a small set of papers, provide evidence that many models being used in explainable AI research are not building on current scientific understanding of explanation. Further, human behavioural experiments are rare --- something that needs to change for us to produce useful explanatory agents.

It is important to note that we are not interpreting the above observations to say that there is not a lot of excellent research on explainable AI. For example, consider \citeA{ribeiro2016should}, who have done some remarkable work on explaining classifiers, and yet scored `0' on the `Data Driven' criteria. Instead, they have constructed their own understanding of how people evaluate explanations for their particular field over a series of  human behavioural experiments. However, developing such an understanding will not always be required or even possible for many researchers, so in these cases, building on social science research is a sound place to start.


\section{Where to? A Brief Pointer to Relevant Work}
\label{sec:where-to}

In the different sub-fields of social sciences, there are several hundred articles on explanation, not to mention another entire field on causality. It is not feasible to expect that AI researchers and practitioners can navigate this entire field in addition to their own field of expertise, especially considering that the relevant literature is written for a different audience. However, there are some key areas that should be of interest to those in explainable AI, which we outline in this section.

\citeA{miller2017explanation-review} provides an in-depth survey of all articles cited in this section plus many other relevant articles, and draws parallels between this work and explainable AI. Here, we present several key ideas from that work  to demonstrate ways that models of explainable AI can benefit from models of  human explanation.

\subsection{Contrastive Explanation}

Perhaps the most important result from this work is that explanations are \emph{contrastive}; or more accurately, \emph{why--questions} are contrastive. That is, why--questions are of the form ``\emph{Why \emph{P} rather than \emph{Q}?}'', where \emph{P} is the \emph{fact} that requires explanation, and \emph{Q} is some \emph{foil} case that was expected. Most philosophers, psychologists, and cognitive scientists in this field assert that \emph{all} why--questions are contrastive (e.g.\ see \cite{hilton1990conversational,lombrozo2012explanation,miller2017explanation-review}), and that when people ask for an explanation ``\emph{Why \emph{P}?}'', there is an implicit contrast case. Importantly, the contrast case helps to frame the possible answers and make them relevant \cite{hilton1990conversational}. For example, explaining ``\emph{Why did Mr.\ Jones open the window?}'' with the response ``\emph{Because he was hot}'' is not useful if the implied foil is Mr.\ Jones turning on the air conditioner, as this explains both the fact and the foil; or if the implied foil was why Ms.\ Smith, who was sitting closer to the window,  did not open it instead, as the cited cause does not refer to a cause of Ms.\ Smith's lack of action.

This is a challenge for explainable AI, because it may not be easy to elicit a contrast case from an observer. However, it is also an opportunity: as \citeA{lipton1990contrastive} argues, answering a contrastive question is often easier than giving a full cause attribution because one only needs to understand the difference between the two cases, so one can provide a complete explanation without determining or even knowing all causes of the event.

\subsection{Attribution Theory}

Attribution theory is the study of how people attribute causes to events; something that is necessary to provide explanations. It is common to divide the types of attribution into two classes: (1) causal attribution of social behaviour (called \emph{social attribution}); and (2) general causal attribution.

\textbf{Social Attribution}~~
The book from \citeA{malle2004mind}, based on a large body of work from himself and other researchers in the field, describes a mature model of how people explain behaviour of others using folk psychology. He argues that people attribute behaviour based on the beliefs, desires, intentions, and traits of people, and presents theories for why failed actions are described differently than successful actions; the former often referring to some precondition that could not be satisfied. 

\citeauthor{malle2004mind}'s work provides a solid foundation on which to build social attribution and explainable AI models for many sub-fields of artificial intelligence. Social attribution is important for systems in which \emph{intentional action} will be cited as a cause; in particular, it is important for systems doing deliberative reasoning, and the concepts used in his work  are closely linked to that of systems such as \emph{belief-desire-intention} models \cite{rao1995bdi} and AI planning.

\textbf{Causal Connection}~~
Research on how people connect causes shows that they do so by undertaking a mental simulation of what \emph{would have happened} had some other event turned out differently \cite{kahneman1982simulation,hilton2005course,mccloy2000counterfactual}.

 However, simulating an entire causal chain is infeasible in most cases, so cognitive scientists and social psychologists have studied how people decide which events to `undo' (the counterfactuals) to determine cause.  For example, people tend to undo more proximal causes over more distal causes \cite{miller1990temporal}, abnormal events over normal events \cite{kahneman1982simulation}, and events that are considered more `controllable' \cite{girotto1991event}.

For explainable AI models, these heuristics are useful from a computational perspective in large causal chains, in which causal attribution is intractable in many cases \cite{eiter2002complexity}. Effectively, they can be used to `skip-over' or \emph{discount} some events and not consider their counterfactuals, while being consistent with what an explainee would expect.

\subsection{Explanation Selection}
\label{sec:where-to:explanation-selection}

An important body of work is concerned with explanation \emph{selection}. People rarely expect an explanation that consists of an actual and complete cause of an event. Instead, explainers select one or two causes and present these as \emph{the} explanation. Explainees are typically able to `fill in' their own causal understanding from just these. Thus, some causes are better explanations than others: events that are `closer' to the fact in question in the causal chain are preferred over more distal events \cite{miller1990temporal}, but people will `trace through' closer events to more distal events if those distal events are human actions \cite{hilton2005course} or abnormal events \cite{hilton1986knowledge}.

In AI, perhaps some models are simple enough that explanation selection would not be valuable, or visualisation would provide a powerful medium to show many causes at once. However, for causal chains with than a handful of causes, we argue that explanation selection can be used to simplify and/or prioritise explanations.

\subsection{Explanation Evaluation}
The work discussed in this section so far looks at how explainees generate and select explanations. There is also a body of work that studies how people evaluate the quality of explanations provided to them. The most important finding from this work is that the probability that the cited cause is actually true is not the most important criteria people use \cite{hilton1996mental}. Instead, people judge explanations based on so-called \emph{pragmatic influences} of causes, which include criteria such as usefulness, relevance, etc.\ \cite{slugoski1993attribution}.

Recent work shows that people prefer explanations that are \emph{simpler} (cite few causes) \cite{lombrozo2007simplicity}, more \emph{general} (they explain more events) \cite{lombrozo2007simplicity}, and \emph{coherent} (consistent with prior knowledge) \cite{thagard1989explanatory}. In particular, \citeA{lombrozo2007simplicity} shows that the people disproportionately prefer simpler explanations over more likely explanations.

These criteria are important to any work in explainable AI. Giving simpler explanations that increase the likelihood that the observer both \emph{understands} and \emph{accepts} the explanation may be more useful to establish trust, if this is the primary goal of the explanation.  Learning from these and adding them as objective criteria to models of explainable AI is important.

\subsection{Explanation as Conversation}

Finally, it is important to remember that explanations are interactive conversations, and that people typically abide by certain rules of conversation \cite{hilton1990conversational}. \emph{Grice's maxims} \cite{grice1975logic}  are the most well-known and widely accepted rules of conversation. In short, they say that in a conversation, people consider the following: (a) quality; (b) quantity; (c) relation; and (d) manner. Coarsely, these respectively mean: only say what you believe; only say as much as is necessary; only say what is relevant; and say it in a nice way. \citeA{hilton1990conversational} argues that as explanations are conversations, they follow these maxims. There is body of research that demonstrates people do follow these maxims, as discussed by \citeA{miller2017explanation-review}.

Note that we are not arguing that explanations must be text or verbal. However, explanations presented in a visual way, for example, should have similar properties, and these maxims offer a useful set of objective criteria. 

\subsection{Where not to go}

Finally, we discuss work that we believe should be discounted in explainable AI. Specifically, two well-known theories of explanation, sometimes cited and used in explainable AI articles, are the \emph{logically deductive model} of explanation \cite{hempel1948studies}, and the \emph{co-variation model} \cite{kelley1967attribution}; both of which have had significant impact.  However, since its publication, researchers found that the logically-deductive model was inconsistent in many ways, and instead derived new models of explanation. Similarly, the co-variation model was found be problematic and did not account for many facets of human explanation \cite{malle2011time}, so was refined into other models, such as those of abnormality described in Section~\ref{sec:where-to:explanation-selection}.

While these models are still cited as part of the history of research in explanation, they are no longer considered valid models of human explanation in cognitive and social science.
We contend, therefore, that explainable AI models should build on these newer models, which are widely accepted, rather than these earlier models.


\section{Conclusions}
\label{sec:conc}

We argued that existing models of how people generate, select, present, and evaluate explanations are highly relevant to explainable AI. Via a brief survey of articles, we provide evidence that little research on explainable AI draws on such models. Although the survey was limited, it is clear from our readings that the observation holds more generally. We pointed to a handful of key articles that we believe could be important, but for a proper presentation and discussion of these, see \citeA{miller2017explanation-review}. 

We encourage researchers and practitioners in explainable AI to collaborate with researchers and practitioners from the social and behavioural sciences, to inform both model design and human behavioural experiments. We do not advocate that every paper on explainable AI should be accompanied by human behavioural experiments ---  proxy studies are valid ways to evaluate models of explanation, especially those in early development, and computational problems are also of interest. However, we support the emphasis in the recent DARPA solicitation~\cite{DARPA2016} on reaching 
``human-in-the-loop 
techniques that developers can use ... for
more intensive human evaluations,' and agree with \citeA{velez2017towards} that to have a real-world impact, ``it is essential that we as a community respect the time and effort involved to do such evaluations.''

We hope that readers of this paper and participants in the workshop agree with our position and, where feasible, adopt existing models and methods  to reduce the risk that it is only the inmates that are running the asylum.


\bibliographystyle{ijcai2017/named}
\bibliography{explanation,other}

\begin{extended}
\pagebreak

\appendix

\section{List of Papers Surveyed}
\label{appendix:paper-list}

Taken from the `Related Work' list posted on the website for the IJCAI 2017 Explainable AI workshop\footnote{See \url{http://home.earthlink.net/~dwaha/research/meetings/ijcai17-xai/}.} as of 16 May 2017.

\begin{enumerate}

\item Chakraborti, T., Sreedharan, S., Zhang, Y., \& Kambhampati, S. (2017). Plan explanations as model reconciliation: Moving beyond explanation as soliloquy. To appear in Proceedings of the Twenty-Sixth International Joint Conference on Artificial Intelligence. Melbourne, Australia: AAAI Press.

\item Cheng, H., et al. (2014) SRI-Sarnoff Aurora at TRECVid 2014: Multimedia event detection and recounting.

\item Doshi-Velez, F., \& Kim, B. (2017). A roadmap for a rigorous science of interpretability. (arXiv:1702.08608)

\item Elhoseiny, M., Liu, J., Cheng, H., Sawhney, H., \& Elgammal, A. (2015). Zero-shot event detection by multimodal distributional semantic embedding of videos. Proceedings of the Thirtieth AAAI Conference on Artificial Intelligence (pp. 3478-3486). Phoenix, AZ: AAAI Press.

\item Hendricks, L.A, Akata, Z., Rohrbach, M., Donahue, J., Schiele, B., \& Darrell, T. (2016). Generating visual explanations. (arXiv:1603.08507v1)

\item
Kofod-Petersen, A., Cassens, J., \& Aamodt, A. (2008). Explanatory capabilities in the CREEK knowledge-intensive case-based reasoner. Frontiers in Artificial Intelligence and Applications, 173, 28-35.

\item
Kulesza, T., Burnett, M., Wong, W. K., \& Stumpf, S. (2015). Principles of explanatory debugging to personalize interactive machine learning. Proceedings of the Twentieth International Conference on Intelligent User Interfaces (pp. 126-137). Atlanta, GA: ACM Press.

\item
Lake, B.H., Salakhutdinov, R., \& Tenenbaum, J.B. (2015). Human-level concept learning through probabilistic program induction. Science, 350, 1332-1338.

\item
Langley, P., Meadows, B., Sridharan, M., \& Choi, D. (2017). Explainable agency for intelligent autonomous systems. In Proceedings of the Twenty-Ninth Annual Conference on Innovative Applications of Artificial Intelligence. San Francisco: AAAI Press.

\item L\'{e}cu\'{e}, F. (2012). Diagnosing changes in an ontology stream: A DL reasoning approach. In Proceedings of the Twenty-Sixth AAAI Conference on Artificial Intelligence. Toronto, Ontario, Canada: AAAI Press.

\item Letham, B., Rudin. C., McCormick, T., and Madigan, D. (2015). Interpretable classifiers using rules and Bayesian analysis: Building a better stroke prediction model. Annals of Applied Statistics, 9(3), 1350-137.

\item Lombrozo, T. (2012). Explanation and abductive inference. Oxford Handbook of Thinking And Reasoning (pp. 260-276).

\item Martens, D., \& Provost, F. (2014). Explaining data-driven document classifications. MIS Quarterly, 38(1), 73-99.

\item Ribeiro, M.T., Singh, S., \& Guestrin, C. (2016). "Why should I trust you?" Explaining the predictions of any classifier. Human Centered Machine Learning: Papers from the CHI Workshop. (arXiv:1602.04938v1)

\item Rosenthal, S., Selvaraj, S. P., \& Veloso, M. (2016). Verbalization: Narration of autonomous mobile robot experience. In Proceedings of the Twenty-Fifth International Joint Conference on Artificial Intelligence. New York, NY: AAAI Press.

\item  Sheh, R.K. (2017). ``Why did you do that?'' Explainable intelligent robots. In K. Talamadupula, S. Sohrabi, L. Michael, \& B. Srivastava (Eds.) Human-Aware Artificial Intelligence: Papers from the AAAI Workshop (Technical Report WS-17-11). San Francisco, CA: AAAI Press.

\item Si, Z. and Zhu, S. (2013). Learning AND-OR templates for object recognition and detection. IEEE Transactions On Pattern Analysis and Machine Intelligence, 35(9), 2189-2205.

\item Shwartz-Ziv, R. \& Tishby, N. (2017). Opening the black box of deep neural networks via information. (arXiv:1703.00810 [cs.LG])

\item Sormo, F., Cassens, J., \& Aamodt, A. (2005). Explanation in case-based reasoning: Perspectives and goals. Artificial Intelligence Review, 24(2), 109-143.

\item
Swartout, W., Paris, C., \& Moore, J. (1991). Explanations in knowledge systems: Design for explainable expert systems. IEEE Expert, 6(3), 58-64.

\item
van Lent, M., Fisher, W., \& Mancuso, M. (2004). An explainable artificial intelligence system for small-unit tactical behavior. Proceedings of the Nineteenth National Conference on Artificial Intelligence (pp. 900-907). San Jose, CA: AAAI Press.

\item 
Zahavy, T., Zrihem, N.B., \& Mannor, S. (2017). Graying the black box: Understanding DQNs. (arXiv:1602.02658 [cs.LG])

\item
Zhang, Y., Sreedharan, S., Kulkarni, A., Chakraborti, T., Zhuo, H.H., \& Kambhampati, S. (2017). Plan explicability and predictability for robot task planning. To appear in Proceedings of the International Conference on Robotics and Automation. Singapore: IEEE Press.

\end{enumerate}

\pagebreak

\section{Detailed Results}
\label{appendix:data}

~

\begin{tabular}{lrrrp{7.5cm}}
\toprule									
\textbf{Paper}	&	\textbf{On topic}	&	\textbf{Data Driven}	&	\textbf{Validation}	&	\textbf{Comments}	\\[1mm] \midrule
1	&	1	&	1	&	0	&		\\[1mm]
2	&	0	&	0	&	0	&		\\[1mm]
3	&	1	&	1	&	N/A	&	A position paper, so Validation not applicable.	\\[1mm]
4	&	0	&	0	&	0	&		\\[1mm]
5	&	1	&	0	&	1	&		\\[1mm]
6	&	1	&	1	&	0	&		\\[1mm]
7	&	1	&	0	&	1	&		\\[1mm]
8	&	0	&	2	&	1	&	Off topic, but is mature work	\\[1mm]
9	&	1	&	0	&	N/A	&		\\[1mm]
10	&	0	&	0	&	0	&		\\[1mm]
11	&	1	&	?	&	0	&	Could not locate reference Jennings et al. (1982)	\\[1mm]
12	&	N/A	&	N/A	&	N/A	&	Survey paper on explanation in the social sciences	\\[1mm]
13	&	1	&	0	&	0	&		\\[1mm]
14	&	1	&	0	&	1	&		\\[1mm]
15	&	1	&	0	&	0	&		\\[1mm]
16	&	1	&	0	&	0	&		\\[1mm]
17	&	0	&	0	&	0	&		\\[1mm]
18	&	1	&	0	&	0	&		\\[1mm]
19	&	1	&	2	&	N/A	&	Survey paper, so Validation not applicable	\\[1mm]
20	&	1	&	0	&	0	&		\\[1mm]
21	&	1	&	0	&	1	&		\\[1mm]
22	&	1	&	0	&	0	&		\\[1mm]
23	&	1	&	1	&	0	&		\\[1mm] \bottomrule
\end{tabular}

\end{extended}

\end{document}